\documentclass[submission,copyright,creativecommons]{eptcs}
\providecommand{\event}{TARK 2023} 

\usepackage{iftex}
\usepackage{multirow}
\usepackage{xcolor}
\usepackage{graphicx}
\usepackage{svg}

\ifpdf
  \usepackage{underscore}         
  \usepackage[T1]{fontenc}        
\else
  \usepackage{breakurl}           
\fi

\title{Humans in Humans Out:\\ On GPT Converging Toward Common Sense in both Success and Failure}
\author{Philipp Koralus
\institute{Faculty of Philosophy\\ University of Oxford}
\institute{Institute for Ethics in AI\\
University of Oxford}
\institute{St. Catherine's College, Oxford}
\email{philipp.koralus@stcatz.ox.ac.uk}
\and
Vincent Wang-Ma\'scianica
\institute{Department of Computer Science\\
University of Oxford}
\institute{St. Catherine's College, Oxford}
\email{vincent.wang@stcatz.ox.ac.uk}
}
\def\titlerunning{Humans in Humans Out: On GPT Converging Toward Common Sense in both Success and Failure}
\def\authorrunning{Philipp Koralus, Vincent Wang-Ma\'scianica}

\begin{document}
\maketitle

\begin{abstract}
Increase in computational scale and fine-tuning has seen a dramatic improvement in the quality of outputs of large language models (LLMs) like GPT \cite{openai-gpt-4-2023,brown-language-2020}. Given that both GPT-3 and GPT-4 were trained on large quantities of human-generated text, we might ask to what extent their outputs reflect patterns of human thinking, both for correct and incorrect cases. The Erotetic Theory of Reason (ETR) provides a symbolic generative model of both human success and failure in thinking, across propositional \cite{10.1093/oso/9780198823766.003.0002}, quantified \cite{10.1093/oso/9780198823766.003.0004}, and probabilistic reasoning \cite{10.1093/oso/9780198823766.003.0005}, as well as decision-making \cite{10.1093/oso/9780198823766.003.0006}. We presented GPT-3, GPT-3.5, and GPT-4 with 61 central inference and judgment problems from a recent book-length presentation of ETR \cite{koralus-reason-2022}, consisting of experimentally verified data-points on human judgment and extrapolated data-points predicted by ETR, with correct inference patterns as well as fallacies and framing effects (the ETR61 benchmark). ETR61 includes classics like Wason's card task, illusory inferences, the decoy effect, and opportunity-cost neglect, among others. GPT-3 showed evidence of ETR-predicted outputs for 59\% of these examples, rising to 77\% in GPT-3.5 and 75\% in GPT-4. Remarkably, the production of human-like fallacious judgments increased from 18\% in GPT-3 to 33\% in GPT-3.5 and 34\% in GPT-4. This suggests that larger and more advanced LLMs may develop a tendency toward more human-like mistakes, as relevant thought patterns are inherent in human-produced training data. According to ETR, the same fundamental patterns are involved both in successful and unsuccessful ordinary reasoning, so that the "bad" cases could paradoxically be learned from the "good" cases. We further present preliminary evidence that ETR-inspired prompt engineering could reduce instances of these mistakes.
\end{abstract}

\section{Introduction}

Large language models (LLMs) like GPT-4 have become increasingly important due to their wide-ranging capabilities and are already having a transformative impact in many applications. LLMs are designed to generate human-like text, as well as other types of outputs, as reasonable responses to prompts. While exact figures for GPT-4 have not been released, GPT-3 was trained on hundreds of billions of words from diverse sources, including websites, books, and articles. Besides very large data-sets of human-generated text, GPT-3.5 and GPT-4 both also benefited from a human-guided fine-tuning training regime \cite{ziegler2020finetuning}. Given that these systems have been trained on human input rather than formal knowledge representations and their outputs are determined by extrapolation of patterns found in those inputs (essentially a form of next-word prediction), rather than by formal deductions, we may ask whether these systems approximate human outputs in \emph{failure}, as well as in success. To give structure to this investigation, we use an empirically motivated model of human reason, the erotetic theory.

\subsection{Success and failure in human reason through the lens of the erotetic theory}

Human reason has a two-faced nature. On the one hand, it has an unparalleled range of successes to its credit, notably science and modern civilisation. On the other hand, humans suffer from a variety of systematic failures of reason. Many such experimentally documented failures have been widely popularised in the literature \cite{Kahneman:2011, ariely-predictably-2009}. According to the erotetic (or ``question-based'') theory of reason (ETR), the two-faced nature of human reason is the result of one set of basic principles. We treat incoming information, as well as choice scenarios, as raising questions and we aim to resolve these questions as directly as possible. In many situations, the dynamics of resolving questions as directly as possible yields correct judgments, but it can also lead us astray. Luckily, the processes proposed by ETR are guaranteed to produce correct judgments, by the relevant standards of classical validity, probabilistic coherence, and rational choice theory, provided that the agent is in \emph{erotetic equilibrium}, e.g. in a state from which the judgment could be reached regardless of what further questions might be raised. Unfortunately, such equilibrium is often too costly to achieve. A key tenet of the theory is that common-sense judgment is following the same core principles, regardless of whether it is succeeding or leading us astray; the only difference is whether we are robust with respect to further questions. We will illustrate the basic idea with a simple example.

\begin{itemize}
\item[(1)] You have a hand of several cards. There is at least an ace and a queen in the hand or at least a king and a jack. There is an ace in the hand. What if anything follows?
\end{itemize}

A majority of participants in a variety of studies of this type of pattern freely conclude that it follows that there is a queen in the hand \cite{walsh-coreference-2004, koralus-illusory-2018}. A moment's reflection shows that this is not the case: the premises are compatible with a situation in which there is an ace, no queen, but a king and a jack. The erotetic theory diagnoses the human tendency toward this fallacy as follows: we take the premise on board as if it was a question, \texttt{``Am I in an ace and queen situation, or in a king and jack situation?''} We then treat the next premise as a maximally strong answer, \texttt{``You're in an ace situation!''} Your ace situation has a queen in it as well, so you conclude that there is a queen. 

ETR holds that the same question/answer dynamics are behind ordinary judgment more broadly, including cases in which we draw correct inferences. For example, consider \emph{modus ponens}. From \texttt{``If there is an ace in the hand, then there is king in the hand''} and \texttt{``There is an ace in the hand''}, virtually everybody readily infers \texttt{``There is a king in the hand''}. ETR also treats this as a case of the same question/answer dynamic we have already considered: we take the conditional premise on board as a question, \texttt{``Am I in an ace and king situation, or am I in a non-ace situation?''} We then take the second premise as, \texttt{``You are in an ace situation!''}, and, again, treat it as a maximally strong answer, yielding the conclusion that there is a king. In other words, the same basic dynamics involved in correct reasoning are also involved in fallacious reasoning. The key difference, according to ETR, is that fallacious judgments are not in equilibrium with respect to further questions that might be asked. Returning to the first question, suppose I have taken on board: \texttt{``Am I in an ace and queen situation, or in a king and jack situation?''} If I now further ask myself, e.g. \texttt{``What about further situations with respect to the question of whether there is an ace?''}, I end up with, \texttt{``Am I in an ace and a queen situation, a king, jack, and ace situation, or a king, jack, and no ace situation?''} If I now take on board \texttt{``You are in an ace situation!''} as a maximally strong answer as before, I no longer get the conclusion that there is a king. The fallacious inference was not an erotetic equilibrium inference. However, if we gave the \emph{modus ponens} inference the same treatment of asking further questions, we would still arrive at the conclusion that there is a king. This is because correct inferences are erotetic equilibrium inferences; they are robust to further questions. 

The patterns of judgment errors that can be explained erotetically are not limited to logical inference. Similar patterns can be identified in reasoning about probability. For example, it is well-known that humans have a tendency in certain cases to rank the probability of a conjunction as higher than the probability of individual conjuncts, violating the axioms of probability \cite{Tversky:1983}. If we are told that someone is a math genius as well as an athletic outdoorswoman, we might na\"ively judge it to be more likely for this person to be a computer scientist and active in the climbing community, than for this person to be active in the climbing community. The ETR diagnosis is that \texttt{``You're in a math genius and athletic outdoorswoman situation!''} more strongly answers in favour of the conjunction (provided no further questions are asked).

To consider another example from the domain of decision-making, it is known that we suffer from irrational opportunity-cost neglect \cite{frederick-opportunity-2009}. If we are told that we have saved some money for fun in a hypothetical scenario, there is a greater chance we will say \texttt{``Buy!''} to \texttt{``Buy an entertaining video or don't buy an entertaining video?''} than if we are asked \texttt{``Buy an entertaining video or save your money for other purchases?''} The ETR diagnosis is that \texttt{``Fun!''} more strongly answers in favour of buying the video than in favour of not buying the video -- unless you remind yourself that not buying the video is a situation that likely has other potential fun purchases down the line, and make that part of what your decision-question considers.

The question/answer dynamic proposed by ETR is represented set-theoretically, treating questions fundamentally as sets of alternative states, with the dynamics of inference and judgment treated as update rules on sets. ETR to-date covers propositional reasoning \cite{10.1093/oso/9780198823766.003.0002, koralus-erotetic-2013}, first-order quantified relational reasoning \cite{10.1093/oso/9780198823766.003.0004}, reasoning with uncertainty \cite{10.1093/oso/9780198823766.003.0005} and decision-making \cite{10.1093/oso/9780198823766.003.0006}. ETR is symbolic, in the sense that its outputs are determined by sets of states and the atoms that occur within those states. ETR is a generative theory, in the sense that it produces concrete inferences from a given sequence of inputs. In other words, we can ask it what if anything follows from a sequence of premises, or what the decision is for a given decision-question and a given a set of priorities. The theory has been motivated by a large number of data-points from the empirical literature on human reasoning and judgment, and has also been corroborated by experiments on ETR-specific predictions. For example, answering a question is an inherently ordered process, since you cannot meaningfully treat something as an answer without having first taken on board a question. ETR correctly predicts that the extent to which people fall for our opening example (1) of a fallacious inference is lessened if the order of premises is reversed \cite{koralus-illusory-2018}. After all, if we are first given \texttt{``You are in an ace situation!''} and then \texttt{``Am I in an ace and queen situation, or in a king and jack situation?''} it is harder to treat the former as an answer to the latter.

\subsection{Predicting failure patterns}

According to ETR, the same dynamics of reason underly common sense thinking in both success and failure. This would lead us to expect that if we were to train a system to approximate the dynamics of common sense thinking based on human-produced text, the familiar fallacious judgment patterns should emerge as well. On the ETR view, this would be the case even if the text used as a training basis vastly over-represents cases of \emph{correct} common-sense judgment. This is relevant since we expect that GPT would have been trained somewhat judiciously to approximate high-quality output and to solve problems with the hope of objective correctness. Yet, on the ETR view, it would not be a a surprise if a better approximation of \emph{good} common-sense reasoning also yields a better approximation of \emph{bad} common-sense reasoning. The ETR view of how the successes and failures of reason are linked contrasts with views that might take mistakes to be the result of special purpose heuristics. In addition, insofar as judgment errors are made as a result of approximating patterns of common-sense judgment as envisaged by ETR, we would further expect that prompts to take on board broader questions should reduce the occurrence of fallacious judgments.

To investigate these hypotheses, we studied the outputs of GPT with key examples from a recent book-length treatment of ETR as inputs. We then examined whether ETR-inspired prompt engineering can reduce the incidence of certain fallacious judgments that the GPT systems produced.

\section{Study 1: Reasoning patterns in GPT with the ETR61 benchmark}

We logged the outputs of GPT-3, GPT-3.5, and GPT-4 for a set of 61 inference and decision-making problems for which ETR makes predictions about common-sense responses. We will refer to this set as the ETR61 benchmark. ETR61 was created by compiling the main examples throughout a book-length presentation of ETR \cite{koralus-reason-2022}, including some novel content variations. The examples covered a variety of types of problems, across logical and probabilistic inference, as well as decision-making. Many examples were variants of cases that have been shown experimentally to elicit incorrect or incoherent responses in humans. Some additional examples are reproduced below.

\begin{itemize}
\item[(2)] Some blue cards are textured. All square cards are blue. What if anything follows?
\end{itemize}

Human participants tend to fallaciously conclude that some square cards are textured \cite{khemlani-theories-2012}.

\begin{itemize}
\item[(3)] There are several cards on the table, which have a letter on one side and a number on the other side. One card shows and E, one card shows a C, one card shows a 4, and one card shows a 5.  Which cards do you have to turn over to determine if the following statement is true? If a card has an E on one side then it has a four on the other side.
\end{itemize}

In this type of problem the majority of human participants omit to turn over the "5" card, even though it is necessary to do so to determine whether the conditional is falsified \cite{Wason1966-WASR}.

\begin{itemize}
\item[(4)] [Abbreviated] Linda is thirty-one years old. She majored in philosophy. As a student, she was deeply concerned with issues of discrimination and social justice. Please rank order by probability (highest to lowest) the following: Linda is a bank teller. Linda is a bank teller and is active in the feminist moment.
\end{itemize}

Human participants have a tendency to rank the probability of the conjunction higher than the probability of the conjuncts \cite{Tversky:1983}, violating the axioms of probability theory.

\begin{itemize}
\item[(5)] Which of the following subscription would you be most likely to purchase. 1. Economist.com subscription - US \$59.00. One-year subscription to Economist.com. Includes online access to all articles from The Economist since 1997. [2. Print subscription - US \$ 125.00. One-year subscription to the print edition of The Economist.] 3. Print \& web subscription - US \$125.00. One-year subscription to the print edition of The Economist and online access to all articles from The Economist since 1997.
\end{itemize}

Human participants have been reported as being more likely to pick option 3 if the dominated option 2 (which nobody chose) is included in the menu \cite{ariely-predictably-2009}, violating the axioms of rational choice theory.

In constructing the ETR61 benchmark, no systematic commitment was made to replicate the vignettes from human experiments exactly (though most of them were exact replications). We changed the wording and parameters of various problems to overcome the resistance of GPT to make recommendations, and to deal with the possibility of some problems having been strongly present in the training data. We only varied examples in ways that left the ETR predictions on those problems unchanged.

\subsection{Methodology}

The GPT-3 model used was accessed at \href{https://platform.openai.com/playground}{\texttt{https://platform.openai.com/playground}}, with parameters text-davinci-003 at temperature 0.7, max-output 256 tokens, top P = 1, 0 frequency  and presence penalities, best-of-1. We refer to ChatGPT as GPT-3.5 and ChatGPT with GPT-4 backend as just GPT-4. Both GPT-3.5 and GPT-4 models used were accessed at \href{https://chat.openai.com/chat}{\texttt{https://chat.openai.com/chat}} in March 2023, using the default 3.5 and 4 models.

For each version of GPT, we used ETR61 as prompts under two conditions: production and query. Each question in each condition was entered into a refreshed context-free instance of the model.  In the production condition, questions were asked in the form of premises ended by \texttt{``What, if anything, follows?"}. Answers that that did not mention the main ETR-predicted conclusion but included statements that imply it were further prompted with a topical query. For GPT-3, which had a relatively limited max-output length, submission was repeated whenever answers were cut midway.

In the query condition, questions were asked in the form of premises ended by \texttt{``Does it follow that X?"} in place of \texttt{``What, if anything follows?"} where $X$ was the ETR-predicted response. No follow-up questions were applied in the query condition. For both cases, appropriate semantic variants of \texttt{``What, if anything follows?"} or \texttt{``Does it follow that X?"} were assigned for questions involving probability rankings and decision-making.

For every question in both conditions, correctness was recorded and the transcript logged. Correctness was judged by the following criteria. A correct declaration of the answer anywhere was marked correct regardless of the ensuing justification. This gave GPT-3 an inflated correctness rate, as it often produced nonsense justifications. Mindful of poor native numerical ability in language models \cite{frieder-mathematical-2023} (and workarounds \cite{wolfram-chatgpt-2023}), for questions concerning probability, correctness was awarded for either correct answers or correct procedure \emph{e.g.} correct application of Bayes' rule. For rankings of subjective probability, correctness was awarded for probabilistic consistency \emph{e.g.} $\mathbf{P}(A) \geq \mathbf{P}(A \& B)$ anywhere is inconsistent. For decision questions, correctness was awarded for consistent decision-making across materially equivalent framings of the choice scenarios.

\subsection{Results}

We adopt the convention of statistical significance at $p \leq 0.05$. All significance tests are Wilcoxon signed-rank as implemented in Mathematica version 13.0.1.0. by the SignedRankTest method. Since it is known \cite{bergmann-different-2000} that different computer packages for rank-sum statistics may yield different judgements of significance at the $p = 0.05$ level, we report fair significance ($0.01 \leq p \leq 0.05$) in \textcolor{orange}{orange} and strong significance ($p \leq 0.01$) in \textcolor{red}{red}. We introduce each test condition with a gloss, we report the percentage rate for each test across the GPTs, and we provide the statistical significance of the change in rates; we tabulate the change in rate from GPT-3 to GPT-3.5 as ``v3-v3.5", GPT-3.5 to GPT-4 as ``v3.5-v4", and GPT-3 to GPT-4 as ``v3-v4".

We found that GPT-3.5 produced fewer correct answers than either GPT-3 or GPT-4. In Table \ref{table:1} we observe a dip-then-rise in the rate of production of correct answers across generations, and a steady-then-rise in the endorsement of correct answers. We observe a steady-then-sharp-rise in correctness and consistency, taken to be both the production and endorsement of correct answers.

\begin{table}[ht!]
\centering
\begin{tabular}{|ccccc|}
\hline
\multicolumn{5}{|c|}{\emph{\textbf{How correct are GPTs as productive reasoners in ETR61?}}}   \\ \hline
\multicolumn{1}{|c|}{\multirow{4}{*}{Correct answer produced}}
& \multicolumn{1}{c|}{\multirow{2}{*}{Percentage}}
&\multicolumn{1}{c|}{GPT-3} & \multicolumn{1}{c|}{GPT-3.5} & GPT-4 \\ \cline{3-5} 
\multicolumn{1}{|c|}{} & \multicolumn{1}{c|}{}&

\multicolumn{1}{c|}{\textbf{56\%}} &
\multicolumn{1}{c|}{\textbf{48\%}}
& \textbf{64\%}           

\\ \cline{2-5} 
\multicolumn{1}{|c|}{} & \multicolumn{1}{c|}{\multirow{2}{*}{\begin{tabular}[c]{@{}c@{}}Cross-generational\\ statistical significance\end{tabular}}} & \multicolumn{1}{c|}{v3-v3.5} & \multicolumn{1}{c|}{v3.5-v4}   & v3-v4 \\ \cline{3-5} 
\multicolumn{1}{|c|}{}                      & \multicolumn{1}{c|}{} 

& \multicolumn{1}{c|}{\textbf{0.3248}} 
& \multicolumn{1}{c|}{\textbf{\textcolor{orange}{0.0346}}} 
& \textbf{0.3752}    
\\ \hline

\multicolumn{5}{|c|}{\emph{\textbf{How often do GPTs endorse correct conclusions in ETR61?}}}   \\ \hline
\multicolumn{1}{|c|}{\multirow{4}{*}{Correct answer endorsed}}
& \multicolumn{1}{c|}{\multirow{2}{*}{Percentage}}
&\multicolumn{1}{c|}{GPT-3} & \multicolumn{1}{c|}{GPT-3.5} & GPT-4 \\ \cline{3-5} 
\multicolumn{1}{|c|}{} & \multicolumn{1}{c|}{}&

\multicolumn{1}{c|}{\textbf{48\%}} &
\multicolumn{1}{c|}{\textbf{52\%}}
& \textbf{72\%}           

\\ \cline{2-5} 
\multicolumn{1}{|c|}{} & \multicolumn{1}{c|}{\multirow{2}{*}{\begin{tabular}[c]{@{}c@{}}Cross-generational\\ statistical significance\end{tabular}}} & \multicolumn{1}{c|}{v3-v3.5} & \multicolumn{1}{c|}{v3.5-v4}   & v3-v4 \\ \cline{3-5} 
\multicolumn{1}{|c|}{}                      & \multicolumn{1}{c|}{} 

& \multicolumn{1}{c|}{\textbf{0.6701}} 
& \multicolumn{1}{c|}{\textbf{\textcolor{orange}{0.0150}}} 
& \textbf{\textcolor{red}{0.0045}}    
\\ \hline

\multicolumn{5}{|c|}{\emph{\textbf{How often is GPT consistently correct in ETR61?}}}   \\ \hline
\multicolumn{1}{|c|}{\multirow{4}{*}{Correct production and endorsement}}
& \multicolumn{1}{c|}{\multirow{2}{*}{Percentage}}
&\multicolumn{1}{c|}{GPT-3} & \multicolumn{1}{c|}{GPT-3.5} & GPT-4 \\ \cline{3-5} 
\multicolumn{1}{|c|}{} & \multicolumn{1}{c|}{}&

\multicolumn{1}{c|}{\textbf{30\%}} &
\multicolumn{1}{c|}{\textbf{30\%}}
& \textbf{59\%}           

\\ \cline{2-5} 
\multicolumn{1}{|c|}{} & \multicolumn{1}{c|}{\multirow{2}{*}{\begin{tabular}[c]{@{}c@{}}Cross-generational\\ statistical significance\end{tabular}}} & \multicolumn{1}{c|}{v3-v3.5} & \multicolumn{1}{c|}{v3.5-v4}   & v3-v4 \\ \cline{3-5} 
\multicolumn{1}{|c|}{}                      & \multicolumn{1}{c|}{} 

& \multicolumn{1}{c|}{\textbf{1}} 
& \multicolumn{1}{c|}{\textbf{\textcolor{red}{0.0004}}} 
& \textbf{\textcolor{red}{0.0017}}    
\\ \hline
\end{tabular}
\caption{Cross-generational tests for correctness in reasoning across GPTs.}
\label{table:1}
\end{table}

\emph{More advanced models exhibit more common sense.} In Table \ref{table:2} we observe upward trends in rates of answers produced and answers endorsed that are predicted by ETR as common-sense judgments. There is a rise-then-steady pattern when we consider any instance of common-sense reasoning in either production or endorsement. While these trends are not statistically significant stepwise between generations, there is strong evidence that GPT-4 is overall more common-sense congruent than GPT-3.

\begin{table}[ht!]
\centering
\begin{tabular}{|ccccc|}
\hline
\multicolumn{5}{|c|}{\emph{\textbf{How often do GPTs produce ETR-predicted common-sense answers?}}}   \\ \hline
\multicolumn{1}{|c|}{\multirow{4}{*}{Production predicted by ETR}}
& \multicolumn{1}{c|}{\multirow{2}{*}{Percentage}}
&\multicolumn{1}{c|}{GPT-3} & \multicolumn{1}{c|}{GPT-3.5} & GPT-4 \\ \cline{3-5} 
\multicolumn{1}{|c|}{} & \multicolumn{1}{c|}{}&

\multicolumn{1}{c|}{\textbf{49\%}} &
\multicolumn{1}{c|}{\textbf{61\%}}
& \textbf{72\%}           

\\ \cline{2-5} 
\multicolumn{1}{|c|}{} & \multicolumn{1}{c|}{\multirow{2}{*}{\begin{tabular}[c]{@{}c@{}}Cross-generational\\ statistical significance\end{tabular}}} & \multicolumn{1}{c|}{v3-v3.5} & \multicolumn{1}{c|}{v3.5-v4}   & v3-v4 \\ \cline{3-5} 
\multicolumn{1}{|c|}{}                      & \multicolumn{1}{c|}{} 

& \multicolumn{1}{c|}{\textbf{0.1134}} 
& \multicolumn{1}{c|}{\textbf{0.1316}} 
& \textbf{\textcolor{red}{0.0030}}    
\\ \hline

\multicolumn{5}{|c|}{\emph{\textbf{How often do GPTs endorse ETR-predicted common-sense answers?}}}   \\ \hline
\multicolumn{1}{|c|}{\multirow{4}{*}{Endorsement predicted by ETR}}
& \multicolumn{1}{c|}{\multirow{2}{*}{Percentage}}
&\multicolumn{1}{c|}{GPT-3} & \multicolumn{1}{c|}{GPT-3.5} & GPT-4 \\ \cline{3-5} 
\multicolumn{1}{|c|}{} & \multicolumn{1}{c|}{}&

\multicolumn{1}{c|}{\textbf{36\%}} &
\multicolumn{1}{c|}{\textbf{49\%}}
& \textbf{54\%}           

\\ \cline{2-5} 
\multicolumn{1}{|c|}{} & \multicolumn{1}{c|}{\multirow{2}{*}{\begin{tabular}[c]{@{}c@{}}Cross-generational\\ statistical significance\end{tabular}}} & \multicolumn{1}{c|}{v3-v3.5} & \multicolumn{1}{c|}{v3.5-v4}   & v3-v4 \\ \cline{3-5} 
\multicolumn{1}{|c|}{}                      & \multicolumn{1}{c|}{} 

& \multicolumn{1}{c|}{\textbf{0.1060}} 
& \multicolumn{1}{c|}{\textbf{0.5255}} 
& \textbf{\textcolor{orange}{0.0354}}    
\\ \hline

\multicolumn{5}{|c|}{\emph{\textbf{How often do GPTs exhibit ETR-predicted common-sense responses?}}}   \\ \hline
\multicolumn{1}{|c|}{\multirow{4}{*}{Either above predicted by ETR}}
& \multicolumn{1}{c|}{\multirow{2}{*}{Percentage}}
&\multicolumn{1}{c|}{GPT-3} & \multicolumn{1}{c|}{GPT-3.5} & GPT-4 \\ \cline{3-5} 
\multicolumn{1}{|c|}{} & \multicolumn{1}{c|}{}&

\multicolumn{1}{c|}{\textbf{59\%}} &
\multicolumn{1}{c|}{\textbf{77\%}}
& \textbf{75\%}           

\\ \cline{2-5} 
\multicolumn{1}{|c|}{} & \multicolumn{1}{c|}{\multirow{2}{*}{\begin{tabular}[c]{@{}c@{}}Cross-generational\\ statistical significance\end{tabular}}} & \multicolumn{1}{c|}{v3-v3.5} & \multicolumn{1}{c|}{v3.5-v4}   & v3-v4 \\ \cline{3-5} 
\multicolumn{1}{|c|}{}                      & \multicolumn{1}{c|}{} 

& \multicolumn{1}{c|}{\textbf{\textcolor{red}{0.0083}}} 
& \multicolumn{1}{c|}{\textbf{0.8213}} 
& \textbf{\textcolor{orange}{0.0268}}    
\\ \hline

\end{tabular}
\caption{Cross-generational tests for common-sense reasoning across GPTs.}
\label{table:2}
\end{table}

\emph{More advanced models are more prone to ETR-predicted fallacies.} We consider fallacies to be "mistakes of common sense" that correspond to incorrect answers that are predicted by the default reasoning procedure in ETR. We note in Table \ref{table:3} that production and endorsement of fallacies rises from GPT-3 to GPT-3.5, and either holds steady or drops slightly from GPT-3.5 to GPT-4. Remarkably, in every case, the more advanced GPT-4 is more fallacy-prone than GPT-3.

\begin{table}[ht!]
\centering
\begin{tabular}{|ccccc|}
\hline
\multicolumn{5}{|c|}{\emph{\textbf{How often do GPTs produce ETR-predicted human-like errors?}}}   \\ \hline
\multicolumn{1}{|c|}{\multirow{4}{*}{Production fallacious}}
& \multicolumn{1}{c|}{\multirow{2}{*}{Percentage}}
&\multicolumn{1}{c|}{GPT-3} & \multicolumn{1}{c|}{GPT-3.5} & GPT-4 \\ \cline{3-5} 
\multicolumn{1}{|c|}{} & \multicolumn{1}{c|}{}&

\multicolumn{1}{c|}{\textbf{18\%}} &
\multicolumn{1}{c|}{\textbf{33\%}}
& \textbf{34\%}           

\\ \cline{2-5} 
\multicolumn{1}{|c|}{} & \multicolumn{1}{c|}{\multirow{2}{*}{\begin{tabular}[c]{@{}c@{}}Cross-generational\\ statistical significance\end{tabular}}} & \multicolumn{1}{c|}{v3-v3.5} & \multicolumn{1}{c|}{v3.5-v4}   & v3-v4 \\ \cline{3-5} 
\multicolumn{1}{|c|}{}                      & \multicolumn{1}{c|}{} 

& \multicolumn{1}{c|}{\textbf{\textcolor{orange}{0.0140}}} 
& \multicolumn{1}{c|}{\textbf{0.8121}} 
& \textbf{\textcolor{orange}{0.0135}}    
\\ \hline

\multicolumn{5}{|c|}{\emph{\textbf{How often do GPTs endorse ETR-predicted human-like errors?}}}   \\ \hline
\multicolumn{1}{|c|}{\multirow{4}{*}{Fallacy endorsed}}
& \multicolumn{1}{c|}{\multirow{2}{*}{Percentage}}
&\multicolumn{1}{c|}{GPT-3} & \multicolumn{1}{c|}{GPT-3.5} & GPT-4 \\ \cline{3-5} 
\multicolumn{1}{|c|}{} & \multicolumn{1}{c|}{}&

\multicolumn{1}{c|}{\textbf{18\%}} &
\multicolumn{1}{c|}{\textbf{23\%}}
& \textbf{20\%}           

\\ \cline{2-5} 
\multicolumn{1}{|c|}{} & \multicolumn{1}{c|}{\multirow{2}{*}{\begin{tabular}[c]{@{}c@{}}Cross-generational\\ statistical significance\end{tabular}}} & \multicolumn{1}{c|}{v3-v3.5} & \multicolumn{1}{c|}{v3.5-v4}   & v3-v4 \\ \cline{3-5} 
\multicolumn{1}{|c|}{}                      & \multicolumn{1}{c|}{} 

& \multicolumn{1}{c|}{\textbf{0.4281}} 
& \multicolumn{1}{c|}{\textbf{0.5941}} 
& \textbf{0.8213}    
\\ \hline

\multicolumn{5}{|c|}{\emph{\textbf{How often do GPTs fall prey to ETR-predicted human-like errors?}}}   \\ \hline
\multicolumn{1}{|c|}{\multirow{4}{*}{Fallacy produced or endorsed}}
& \multicolumn{1}{c|}{\multirow{2}{*}{Percentage}}
&\multicolumn{1}{c|}{GPT-3} & \multicolumn{1}{c|}{GPT-3.5} & GPT-4 \\ \cline{3-5} 
\multicolumn{1}{|c|}{} & \multicolumn{1}{c|}{}&

\multicolumn{1}{c|}{\textbf{26\%}} &
\multicolumn{1}{c|}{\textbf{43\%}}
& \textbf{36\%}           

\\ \cline{2-5} 
\multicolumn{1}{|c|}{} & \multicolumn{1}{c|}{\multirow{2}{*}{\begin{tabular}[c]{@{}c@{}}Cross-generational\\ statistical significance\end{tabular}}} & \multicolumn{1}{c|}{v3-v3.5} & \multicolumn{1}{c|}{v3.5-v4}   & v3-v4 \\ \cline{3-5} 
\multicolumn{1}{|c|}{}                      & \multicolumn{1}{c|}{} 

& \multicolumn{1}{c|}{\textbf{\textcolor{orange}{0.0135}}} 
& \multicolumn{1}{c|}{\textbf{0.2669}} 
& \textbf{0.1647}    
\\ \hline
\end{tabular}
\caption{Cross-generational tests for fallacious reasoning across GPTs.}
\label{table:3}
\end{table}

\emph{GPT-4 produces significantly more fallacies than it endorses.} We performed several intra-generational comparisons, reported in Table \ref{table:4}. First, we know on reasoning problems, human participants often produce different results depending on whether they are asked to produce a conclusion or whether they are asked to whether a given conclusion follows \cite{khemlani-theories-2012}. We observe that later generations appeared to produce more fallacies than they endorsed, and we report that GPT-4 produces significantly more fallacies than it endorses.

\begin{table}[ht!]
\centering
\begin{tabular}{|cccc|}
\hline
\multicolumn{4}{|c|}{\textit{\textbf{Do GPTs produce answers other than what they endorse?}}}                                                                                   \\ \hline
\multicolumn{1}{|c|}{\multirow{2}{*}{ETR predicts production vs. ETR predicts endorsement}} & \multicolumn{1}{c|}{GPT-3}       & \multicolumn{1}{c|}{GPT-3.5}     & GPT-4       \\ \cline{2-4} 
\multicolumn{1}{|c|}{}                      & \multicolumn{1}{c|}{\textbf{0.6552}} & \multicolumn{1}{c|}{\textbf{0.3248}} & \textbf{0.1169} \\ \hline

\multicolumn{4}{|c|}{\textit{\textbf{Do GPTs differ in production vs. verification performance?}}}                                                                                   \\ \hline
\multicolumn{1}{|c|}{\multirow{2}{*}{Production correct vs. endorsement correct}} & \multicolumn{1}{c|}{GPT-3}       & \multicolumn{1}{c|}{GPT-3.5}     & GPT-4       \\ \cline{2-4} 
\multicolumn{1}{|c|}{}                      & \multicolumn{1}{c|}{\textbf{0.3429}} & \multicolumn{1}{c|}{\textbf{0.5588}} & \textbf{0.1450} \\ \hline

\multicolumn{4}{|c|}{\textit{\textbf{Is there a difference in rate of production vs. endorsement of fallacies?}}}                                                                                   \\ \hline
\multicolumn{1}{|c|}{\multirow{2}{*}{Fallacy produced vs. fallacy endorsed}} & \multicolumn{1}{c|}{GPT-3}       & \multicolumn{1}{c|}{GPT-3.5}     & GPT-4       \\ \cline{2-4} 
\multicolumn{1}{|c|}{}                      & \multicolumn{1}{c|}{\textbf{1}} & \multicolumn{1}{c|}{\textbf{0.2610}} & \textbf{\textcolor{orange}{0.0134}} \\ \hline
\end{tabular}
\caption{Intra-generational tests across GPTs.}
\label{table:4}
\end{table}

\subsection{Discussion}

We observe that while we see improvements in correct reasoning with later generations of GPT, we also see an increase in the susceptibility to systematic fallacies. Remarkably, GPT-4 did not produce fewer ETR-predicted fallacious conclusions than GPT-3.5; while GPT-3.5 in fact produced more of these fallacious conclusions than GPT-3. This is would be expected if a mixture of more training (including reinforcement learning from human feedback) and increased model-size increase the fidelity with which human common-sense reasoning patterns (as conceived of by ETR) are approximated. Unfortunately, exact details of the model differences have not been made public, making it difficult to theorise more specifically about the mechanisms responsible for the observed differences in model behaviour.

Some of the relevant outputs produced to GPT 4 were in fact reminiscent of the dynamics of reasoning described by ETR. Though these observations have to be interpreted with caution, they are potentially suggestive.
\begin{itemize}
\item[] \emph{Prompt:} Either Jane is kneeling by the fire and she is looking at the TV or else Mark is standing at the window and he is peering into the garden. Jane is kneeling by the fire. What if anything follows?
\item[] \emph{GPT4:} Since Jane is kneeling by the fire, it follows that the first part of the statement is true. [$\dots$] So, what follows is that Jane is kneeling by the fire and looking at the TV.
\end{itemize}

This is reminiscent of the pattern described in section 1.1, according to which entire disjuncts are treated as alternatives in a question, and the subsequent premise is used as a maximally strong answer, effectively selecting the alternative with the greatest overlap, instead of engaging with the logic of disjunction. We also found some evidence of the ETR-predicted order effect discussed in section 1.1: GPT4 produced the relevant fallacious conclusion for \texttt{"There is an ace and a queen, or else a king and a ten. There is a king."} but not for \texttt{"There is a king. There is an ace and a queen, or else a king and a ten."}

\section{Study 2: Prompt engineering to mitigate fallacies}

As in the case of human judgment, the presence of systematic mistakes raises the question of whether we can mitigate these mistakes by following a better reasoning strategy. Adding global instructions like ``reason step-by-step'' can improve the accuracy of LLM outputs \cite{kojima-large-2023}.  According to ETR, mistakes are due to not having taken on board sufficiently broad questions before treating further information as an answer. We examined whether a simplistic version of this idea could improve GPT performance.
\subsection{Method}

For each of GPT-3, 3.5, and 4, we ran augmented prompts based on a subset of ETR61, following a similar procedure that described previously. For each version of GPT, we only used those original prompts that had produced incorrect judgments that were predicted by the basic step in the ETR default reasoning procedure. In other words, we only examined those cases that yielded mistakes and where those mistakes were most directly generated by the question/answer dynamic at the core of ETR.

To construct the augmented prompts we used the following templates:

\begin{description}
    \item[(Control)] Reason step-by-step for the following problem. [Original prompt inserted here]
    \item[(ETR)] Answer the following question according to this procedure: First, list the premises. Second, turn each premise into a question to make a new list of questions; treat questions as possible alternatives. Third, reason step-by-step using both lists, keeping track of alternatives. [Original prompt inserted here]
\end{description}

\subsection{Results}

We found that both the control prompt template and the ETR prompt template reduced the incidence of incorrect judgment outputs. In the case of GPT3.5, the ETR prompt was better than the control prompt at a level that was statistically significant.

\begin{table}[ht!]
\centering
\begin{tabular}{|cccc|}
\hline
\multicolumn{4}{|c|}{\textit{\textbf{Are fallacies blocked by erotetic prompting?}}}                                                                                             \\ \hline
\multicolumn{1}{|c|}{Model}                           & \multicolumn{1}{c|}{GPT-3}       & \multicolumn{1}{c|}{GPT-3.5}     & GPT-4       \\ \hline
\multicolumn{1}{|c|}{Fallacies blocked by control prompt}   & \multicolumn{1}{c|}{\textbf{4/8}} & \multicolumn{1}{c|}{\textbf{4/17}} & \textbf{8/19} \\ \hline
\multicolumn{1}{|c|}{Fallacies blocked by ETR prompt} & \multicolumn{1}{c|}{\textbf{4/8}} & \multicolumn{1}{c|}{\textbf{14/17}} & \textbf{7/19} \\ \hline
\multicolumn{1}{|c|}{Stat. sig.}                      & \multicolumn{1}{c|}{\textbf{1}}          & \multicolumn{1}{c|}{\textbf{\textcolor{red}{0.0045}}}          & \textbf{0.7656}          \\ \hline
\end{tabular}
\caption{Prompting GPT-3.5 to reason erotetically was significantly better at blocking fallacies compared to step-by-step prompting. On other models both prompts had comparable effect on blocking fallacies.}
\label{table:5}
\end{table}

\section{Conclusion and future work}

The results of Study 1 suggest that GPT-3.5 and GPT-4 outputs are significantly more similar to common-sense judgment than GPT-3 outputs. Remarkably, this also applies to patterns of fallacious judgments. In fact, the \emph{production} of fallacious judgments from ETR61, including deductive reasoning, judgments about probability, and decision-making, increases from GPT-3 to GPT-4. On the view of human reasoning proposed by ETR, this is not surprising. ETR holds that both successful and unsuccessful uses of common sense are based on the same principles, so that a better approximation of the dynamics of common-sense, as provided by more training and larger models, would also generate more common-sense fallacies, even if the LLMs are primarily learning from success cases. Study 2 suggests that some of the insights from ETR could inform strategies for prompt engineering to lessen the incidence of common-sense fallacies in LLMs like GPT.

Study 1 bears on the question of how future improvements of LLMs relate to data scale. Given that the later generation LLMs examined were in fact more prone to human-like fallacies, even though they involved more processing on larger data sets, we speculate that the relevant problems are not easily remedied by more generic training on human-generated text, which will have the relevant thought patterns ingrained in it. However, ETR suggests a possible solution. With an implementation of ETR, it would be possible to generate synthetic datasets of arbitrary size and complexity of fallacy-prone problems and their correct solutions that could be used to train against them. Similarly, an implementation of ETR, could be used to construct a dynamic generative benchmark to test LLM performance. This may help further pave the way toward LLMs as \emph{cleaned-up} common-sense approximators.

\nocite{*}
\bibliographystyle{eptcsini}
\bibliography{KoralusWang-HumansinHumansout}

\end{document}